\pdfoutput=1

\documentclass[11pt]{article}
\usepackage{array}
\usepackage{acl}

\usepackage{times}
\usepackage{latexsym}

\usepackage[T1]{fontenc}

\usepackage[utf8]{inputenc}

\usepackage{microtype}
\usepackage{graphicx}
\usepackage[inline]{enumitem}
\usepackage{amsfonts}
\usepackage{multirow}
\usepackage{graphicx}
\usepackage{amssymb}
\usepackage{hyperref}
\usepackage{arydshln}

\usepackage{multicol}
\usepackage{array}
\usepackage{amsmath}

\usepackage{xcolor}

\usepackage{tikz}

\tikzset{
	font=\LARGE,
	line/.style={thick,rounded corners=1mm},
	arrow/.style={line,->,>=stealth},
	layer/.style={draw,thick,rectangle,minimum height=1cm,fill=yellow!20},
	bg_border/.style={draw,dotted,rectangle,minimum height=1cm,fill=lime!20},
	input/.style={draw,thick,circle,fill=blue!20,minimum size=2mm,inner sep=0,outer sep=0},
	vector/.style={draw,thick,rectangle,fill=yellow!20,minimum width=5mm,minimum height=10mm,inner sep=0, outer sep=0},
	sum/.style={draw,circle,minimum size=0mm,inner sep=0,outer sep=0},
	every node/.style={scale=0.45}}

\title{ICDBigBird: A Contextual Embedding Model for  ICD Code Classification}

\author{%
George Michalopoulos$^1$, Michal Malyska$^{2,3}$, Nicola Sahar$^3$\\
{\bf Alexander Wong$^1$,} {\bf Helen Chen$^1$}\\
    University of Waterloo$^1$, University of Toronto$^2$, Semantic Health$^3$\\
\texttt{\{gmichalo,  alexander.wong, helen.chen\}@uwaterloo.ca}\\
 \texttt{michal.malyska@mail.utoronto.ca}
 \\ \texttt{nick@semantichealth.ai}
}

\begin{document}
\maketitle
\begin{abstract}
The International Classification of Diseases (ICD) system is the international standard for classifying diseases and procedures during a healthcare encounter and is widely used for healthcare reporting and management purposes.  Assigning correct codes for  clinical procedures is important for clinical, operational and financial decision-making in healthcare.

Contextual word embedding models have achieved state-of-the-art results in multiple NLP tasks. However, these models have yet to achieve state-of-the-art results in the ICD classification task since one of their main disadvantages   is that they can only process documents that contain a small number of tokens which is rarely the case with real patient notes. In this paper, we introduce ICDBigBird a BigBird-based model which can integrate a Graph Convolutional Network  (GCN), that takes advantage of the relations between ICD codes in order to create `enriched’ representations of their embeddings,  with a BigBird contextual model that can process larger  documents. Our experiments on a real-world clinical dataset demonstrate the effectiveness of our BigBird-based model on the ICD  classification task as it outperforms the previous state-of-the-art models.

\end{abstract}

\section{Introduction}
Real-world data 
in healthcare refers to patient data routinely collected during clinic encounters such as visits and hospitalization.
After each clinical visit, a set of codes representing diagnostic and procedural information are submitted to various  regulatory agencies \cite{Farkas2008AutomaticCO}. 
The International Classification of Diseases (ICD) system  is the most widely used coding system, maintained by the World Health Organization \cite{Avati2018ImprovingPC}. Assigning the most appropriate   codes is an important task in healthcare since erroneous ICD codes could seriously affect the organization's ability to  accurately measure the patient outcome \cite{ji-etal-2020-dilated}.  

Contextual word embedding  models (such  as  ELMo \cite{peters-etal-2018-deep} and BERT \cite{devlin-etal-2019-bert}) have achieved state-of-art results in many NLP tasks. 
 However, recent attempts of using contextual models on the ICD  classification task have failed to achieve  state-of-the-art results  \cite{zhang-etal-2020-bert} mainly due to the fact that they can only process documents that contain a small number of tokens. 
 Advances such as the BigBird model \cite{zaheer2020bigbird} allows contextual models to process long documents, thus reducing the risk of losing information from the original texts. 
 
 In this paper, we present a novel model for the ICD classification task. Specifically:
(i) we are the first, to the best of our knowledge, to propose the combined usage of a Graph Convolutional Network  (based on the normalized point-wise mutual information) and a contextual embedding model for the ICD classification task; 
(ii)  we introduce a novel attention layer on top of a BigBird model which  has the ability to process long documents; and
(iii)  our experiments on a real-world clinical dataset demonstrate the effectiveness of our ICDBigBird model on the ICD  classification task as it outperforms   previous  state-of-the-art models.
 \section{Proposed ICDBigBird Model}

\begin{figure}[h]
        \centering
        \includegraphics[width=\linewidth]{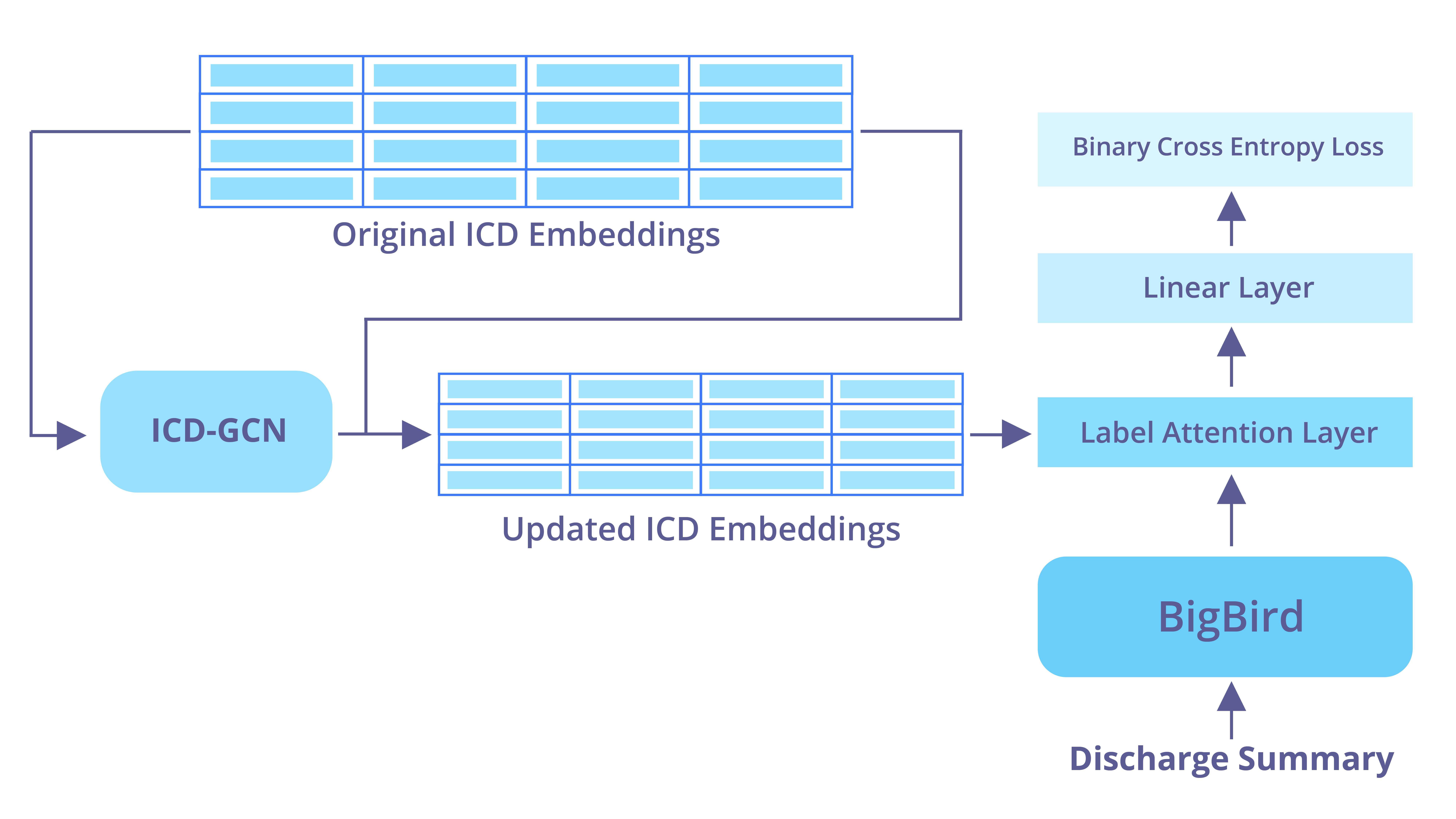}
    \caption{ICDBigBird model architecture}
    \label{fig:1}
\end{figure}

\subsection{ICD Graph Convolutional Network}
\label{methods}
A Graph Convolutional Network (GCN) \cite{kipf2017semisupervised}  is a neural network architecture that can capture the general knowledge about the connections between entities. Specifically, GCN builds a symmetric adjacency matrix based on a predefined relationship graph, and the representation of each node is calculated according to its neighbours. 

We use a GCN to capture a more `enriched' representation for each of the ICD codes. In order to use the ICD-GCN, we first construct the adjacency matrix $A \in \mathbb{R}^{n\times n}$  (where $n$ is the number of unique ICD codes) to represent the connections of ICD codes by using the normalized point-wise mutual information (NPMI) \cite{lu2020vgcn}:

\begin{equation}
NPMI(i,j)= -\frac{1}{\log{p(i,j)}}\log{\frac{p(i,j)}{p(i)p(j)}}
\label{eq:np1}
\end{equation}
where $i$ and  $j$ are different ICD codes and $p(i,j)=\frac{N(i,j)}{N}$, $p(j)=\frac{N(j)}{N}$ and $N(i,j)$ is the number of documents that are labeled with both  $i$ and $j$   codes, $N(i)$ is the number of documents that are labeled with the  $i$   code and  $N$ is the total number of documents of the training set that our model was trained  on. We create an edge between two codes if their NPMI value is greater than a threshold. We empirically set the threshold to $0.2$ by experimenting with different threshold values.

It should be noted that we decided to create   the adjacency matrix of the ICD-GCN by taking advantage of the NPMI values instead of considering the hierarchical associations of the ICD codes because  we mainly focused on the task of classifying the top 50 most frequent  ICD codes \cite{shi2017towards}, where  we found that there exists little to no hierarchical connection between these codes.

We then construct a definition (sentence) embedding matrix for all the ICD codes using their ICD-9 (sentence) definitions from the MIMIC III dataset \cite{mimiciii} and  the pre-trained sentence transformer embedding model in \cite{sentencebert}, which has been shown to outperform other state-of-the-art sentence embedding methods.

An updated representation  of all  ICD codes from the  ICD-GCN is calculated as follows:
\begin{equation}
\hat{U} = Relu(\hat{A}XW)
\label{eq:np2}
\end{equation}
where $X \in \mathbb{R}^{n\times m}$ is the definition embedding matrix,   $n$ is the number of ICD codes,  $m$ is the size of the sentence-definition embedding of each ICD code, $W \in \mathbb{R}^{m\times h}$ is the weight matrix, $h$ is the BigBird's hidden dimension and $\hat{A} =D^{-\frac{1}{2}}AD^{-\frac{1}{2}}$ is the normalized symmetric adjacency matrix where $D_{ii}=\sum_{j}A_{ij}$. 

Finally, we concatenate the output of the ICD-GCN with the initial embeddings  of the ICD codes in order to get a richer representation of the  codes \cite{rios-kavuluru-2018-shot}:
\begin{equation}
U =\hat{U}\mathbin\Vert X  ,  U \in \mathbb{R}^{n\times (m+h)}
\label{eq:np3}
\end{equation}

\begin{table*}
  \begin{tabular}{l c c c c c }
  \hline
   &  \multicolumn{2}{c}{\underline{\hphantom{empoio} \textbf{ AUC-ROC} \hphantom{emptm}}} &  
  \multicolumn{2}{c}{\underline{\hphantom{empt}\textbf{F1}\hphantom{emptyp}}}   \\ 
    Model & Macro   & Micro & Macro & Micro  & $P@5$\\\hline

DRC.\cite{mullenbach-etal-2018-explainable}&88.4&91.6&57.6&63.3 & 61.8\\
LEAM \cite{wang-etal-2018-joint-embedding}&88.1&91.2&54.0&61.9 & 61.2\\

HyperCore \cite{cao-etal-2020-hypercore}& 89.5$\pm$0.3 & 92.9$\pm$ 0.2 &60.9 $\pm$ 0.1 &66.3 $\pm$ 0.1 &63.2 $\pm$  0.2\\

Mult.CNN \cite{lifei}&89.9$\pm$ 0.4&92.8 $\pm$ 0.2&60.6$\pm$1.1&67.0$\pm$0.3 & 64.1$\pm$0.1 \\
DCAN  \cite{ji-etal-2020-dilated}  & \textbf{90.2$\pm$0.6} &\textbf{93.1$\pm$0.1}&61.5$\pm$0.7&67.1$\pm$0.1 & 64.2$\pm$0.2\\
ICDBigBird &90.0$\pm$0.5 & 92.9 $\pm$0.2& \textbf{63.1$\pm$0.5} & \textbf{69.6$\pm$0.1} &  \textbf{65.4$\pm$0.1}\\
ICDBigBird (validation split) &91.0$\pm$0.6 & 93.3 $\pm$0.1& 64.1$\pm$0.4 & 70.4$\pm$0.1 &65.1$\pm$0.3
\\\hdashline
&&\textbf{Ablation Study}&&
\\\hdashline
BERT\cite{devlin-etal-2019-bert} &80.3$\pm$0.4 & 84.4 $\pm$0.5&  43.7$\pm$0.2 &  51.4$\pm$0.5 & 51.9$\pm$0.3  \\ 
BioBERT\cite{10.1093/bioinformatics/btz682} &81.3$\pm$0.5 & 85.5 $\pm$0.4&  46.3$\pm$0.3&  54.6$\pm$0.3 &54.2 $\pm$0.4\\
Bio\_C.\cite{alsentzer-etal-2019-publicly} &81.7$\pm$0.4 & 85.8 $\pm$0.5& 46.4$\pm$0.3&54.3 $\pm$0.4& 53.2$\pm$0.4\\
BigBird (512 tokens) & 80.4$\pm$0.2 & 83.9 $\pm$0.3& 44.9$\pm$0.5 & 52.1$\pm$0.3& 51.2$\pm$0.4\\ 
BigBird (without attention) &86.7$\pm$0.5 & 90.4 $\pm$0.3& 55.2$\pm$0.4 & 64.8$\pm$0.2& 62.5$\pm$0.3\\ 
Linear Attention &88.4$\pm$0.5 & 91.2 $\pm$0.2& 60.2$\pm$0.2 & 67.8$\pm$0.3 &63.6$\pm$0.5 \\ 
R. embedding &89.2$\pm$0.4 & 91.8 $\pm$0.5& 60.8$\pm$0.2 & 67.8$\pm$0.2 & 63.2$\pm$0.1 \\

  \end{tabular}
  \caption{Results of mean $\pm$ standard deviation of  three runs of the ICDBigBird model on the test  split of the MIMIC-III dataset for the top 50 most frequent ICD codes; We also provide the performance of previous state-of-the-art models  using the same test set; \textit{Bio\_C.} is  Bio\_ClinicalBERT; \textit{DRC.} is DR-CAML;  \textit{R. embedding} is a model with  a random initialization of the embeddings of the codes; we also include the results on the validation split of MIMIC III;   Best values on the \textbf{test} set are \textbf{bolded}}
  \label{tab:re}
\end{table*}

\subsection{ICDBigBird Model}
  Assume a  discharge summary has $n$ words, the  model's   tokenizer generates tokens for each word in the document.
  Afterwards the tokens are passed  through multiple attention-based  layers and the model produces the final contextual representation of the document $H \in \mathbb{R}^{t \times h}$ where $t=4096$ is the number of tokens and $h$  is the BigBird's hidden dimension.
We use a fully connected linear layer for the creation of  $\hat{H}$ which is the final embedding representation of the BigBird's embeddings:
\begin{equation}
\hat{H}=   Relu(H W_1) 
\label{eq:np4}
\end{equation}
where $\hat{H} \in \mathbb{R}^{t\times (m+h)} $ and $W_1 \in \mathbb{R}^{h\times (m+h)} $. Afterwards, we apply  a per-label attention mechanism, in order to showcase the most relevant information to the ICD codes in the contextual representation of each document. Formally, using  $U \in \mathbb{R}^{n\times (m+h)} $ which is the `updated' ICD code sentence-definition embedding matrix, we can compute the  attention as:
\begin{equation}
A = SoftMax (U\hat{H}^\top)
\label{eq:np5}
\end{equation}
where $A\in \mathbb{R}^{n\times t} $. After the calculation of the attention score, the output of the attention layer is calculated as:
\begin{equation}
V = A \hat{H}
\label{eq:np6}
\end{equation}
where $V\in \mathbb{R}^{n\times (m+h)}$. Given the `updated' representation $V$, we can compute a probability for each label $l$ by using a pooling operation and a sigmoid transformation over the linear projection of  $V$:
\begin{equation}
\hat{y}= \sigma(pooling(V \circ W))
\label{eq:e7}
\end{equation}
where $W\in \mathbb{R}^{n\times (m+h)} $. 
As the ICD   task is a multi label scenario, the loss function that is typically used   is a multi-label binary cross entropy loss:

\begin{multline}
    \label{loss2}
L_{BCE}(y, \hat{y}) = \sum_{i=1} ^n (y_i log(\hat{y_i})   \\  + (1- y_i) log(1- \hat{y_i}))
\end{multline}
where $y$ is the ground truth label and $\hat{y}$ are the ICD codes that our model predicted for each document. However, due to the extremely imbalance nature of the ICD codes we chose to adopt the Label-Distribution Aware Margin (LDAM) \cite{cao2019learning}. In the LDAM loss function the output value is subtracted by a label-dependent margin $\Delta_i$ before the sigmoid function:
\begin{equation}
\hat{y}' = \sigma(pooling(V \circ W)-\textbf{1}(y_i=1)\Delta_i)
\label{eq:np8}
\end{equation}
where \textbf{1}(.) outputs 1 if $y_i$=1 and $\Delta_i = \frac{C}{n^{1/4}_i}$ where $n_i$ is number of instances of the $i$ ICD code in the training data and $C$ is constant. Thus we use the $L_{LDAM}=L_{BCE}(y, \hat{y}')$.

\section{Experiments}
\subsection{Dataset}

Following previous research work in the ICD classification task \cite{mullenbach-etal-2018-explainable, ji-etal-2020-dilated, lifei}, we conducted our experiments on the subset of the English Multiparameter  Intelligent  Monitoring  in  Intensive  Care III (MIMIC-III) dataset \citep{mimiciii} with the top 50 most frequent  ICD codes \cite{shi2017towards}. Our experiments on this dataset are consistent with its intended use, as it was created and shared for research  purposes (as it stated in its license\footnote{https://tinyurl.com/mimic-licence}). Finally,  we  manually checked  the dataset to investigate the existence of information that  uniquely identifies individual people and offensive content, however, we did not find any indication of either of them.
We extract the free-text discharge summaries and clinical notes, containing  the 50 most frequent  ICD codes, from the MIMIC III dataset   and we concatenate the discharge summaries and notes from the same hospitalization admission  into one single document. 
We use the training/validation/testing split from \cite{mullenbach-etal-2018-explainable,lifei} for a fair comparison. The  document set size of our subset of MIMIC-III is 8066 for training, 1573 for validation and 1729 for testing  respectively. 
Following the prepossessing procedures outlined in \cite{ ji-etal-2020-dilated}, the documents are tokenized and each token is converted to lowercase. Any token that contains no alphabetic characters is removed. 
Instead of truncating the documents to 2500 words, we set the token size limit to 4096 for our ICDBigBird model to take full advantage of the information that can be extracted from each document 
as there are 1345 documents that contain more than 2500 words (with maximum, minimum and average length of 7567, 105, 1609 words respectively).

\subsection{Experimental Setup}

We provide the search strategy and the bound for each hyperparameter as  follows: the batch size is set between 32 and 64, and the learning rate is chosen between the values  $2\mathrm{e}$-$5$, $3\mathrm{e}$-$5$ and $5\mathrm{e}$-$5$. We set the number of training epochs between 25 and 30 epochs to allow for maximal performance. The best values are chosen based on micro-F1 scores\footnote{https://github.com/jamesmullenbach/caml-mimic}  in the validation set. The final hyper-parameters selection of our ICDBigBird model is batch size 32, learning rate $2\mathrm{e}$-$5$, trained on 30 epochs and we empirically set the the C constant of the LDAM loss to 2. We also use the AdamW optimizer  \cite{loshchilov2019decoupled} to optimize the parameters of the model.

All the contextual embedding models are implemented using the transformers library \citep{Wolf2019HuggingFacesTS} on PyTorch 1.7.1. All experiments are executed on a Tesla K80  GPU with 64GB of system RAM on 	Ubuntu 18.04.5 LTS. 

\subsection{Results}
\label{sec:res}
We benchmark our ICDBigBird model against existing state-of-the-art models for the top 50 most frequent ICD classification task. For all models we evaluate the micro and macro averaging F1 score,   the receiver operating characteristic curve (AUC-ROC) and  the precision at k codes with k=5 (P@5). In Table \ref{tab:re}, we can observe that our model outperforms all other models in the micro and macro averaging F1  and in the $P@5$ score with comparable performance on the other two metrics (with the DCAN model \cite{ji-etal-2020-dilated} 
achieving the best AUC-ROC results).  
Finally,   our model contains 110565170 parameters with average running time of   893354 sec.

\subsection{Ablation Study}
\label{sec:ablation}
In order to evaluate the effect of each feature on the performance of ICDBigBird, we conduct an ablation study. The results are presented in Table \ref{tab:re}. 
\begin{enumerate*}[label=(\roman*)]
\item Firstly, we investigate whether the ability of the BigBird model to process large documents can boost the performance of our model. It can be observed that  contextual model architectures that can process small documents of at most 512 tokens (Bert, Biobert, Bio\_ClinicalBert) cannot achieve the  performance of a BigBird architecture even if these models were pre-trained on medical documents (BioBert and Bio\_ClinicalBert). 
\item Furthermore, we examine the performance of the BigBird model when we artificially limit the length of the documents to 512 tokens (BigBird 512 tokens) which is the maximum number of tokens that the BERT model can process. We observe that the performance improvement brought by the BigBird model is lost, making the performance of the BigBird model equivalent to the BERT model. This experiment demonstrates that  one of the main reasons for the BigBird model outperforming the BERT model is the utilization of additional information in larger documents  (4096 tokens) for the ICD automatic encoding task.
\item In addition, we examined the effect of the GCN model by testing the performance of  contextual embeddings without enriching them with  information from the definitions of the codes through an attention mechanism (BigBird without attention) by having an ICD classifier on top of the [cls] token and by substituting the GCN attention mechanism with the typical linear attention mechanism (Linear Attention) \cite{mullenbach-etal-2018-explainable}. It can be observed that our model benefits from the attention mechanism as without it, it cannot achieve optimal performance. 
Also, the fact that the GCN graph attention mechanism    achieves a better performance than a typical linear attention mechanism is a strong indication that the connections between the ICD codes can provide valuable information. 
\item Finally, we investigated the effect of  using the definitions of the codes to initialize their embeddings.  In  our experiments a model with a  random initialization of the embeddings of the codes (R. embedding)  achieved sub-optimal performance and thus we can conclude that using  the codes' definitions  to initialize their embeddings have a positive effect on the model's performance.

\end{enumerate*}

\subsection{Discussion-Related Work}
Recent development in NLP has introduced  deep learning models that can achieve optimal performance on the ICD classification task. In \cite{shi2017towards}, the authors introduced a new model that used word/character embeddings and recurrent neural networks (LSTM) to generate  representations of  the  diagnosis descriptions and of the ICD codes.  In addition, the authors in \cite{mullenbach-etal-2018-explainable} introduced an attention based convolutional neural network (CNN) model which incorporates an attention mechanism in order to identify the most relevant segments that contain medical information.

Furthermore, prior work has explored the use of GCNs for the ICD classification task \cite{rios-kavuluru-2018-shot,chalkidis-etal-2020-empirical} and our attention mechanism can be viewed as an extension of the structured  attention mechanism of \cite{cao-etal-2020-hypercore}. However, some of the  differences between the models are that:
\begin{enumerate*}[label=(\roman*)]
\item Our work uses a  normalized point-wise mutual information  policy to create the edges, while the model in \cite{cao-etal-2020-hypercore} used the co-appearing  values to create a weighted graph. This is a key difference in the ICD coding problem as the method in \cite{cao-etal-2020-hypercore} does not capture the relation between two highly correlated but `unpopular' codes.
\item In addition, the authors in \cite{cao-etal-2020-hypercore} created the code embedding vectors by averaging the word  embeddings of its descriptor, and our work uses pre-trained sentence embedding models which have achieved better performance. 
\item Finally, the model in  \cite{cao-etal-2020-hypercore} used a Convolution Neural Network (CNN) encoder while our work used a  contextual  (BigBird) model to produce   document embeddings.
         \end{enumerate*}     

The results  of the experiments  
indicate that these changes  are important for the ICD  classification task by demonstrating    that a contextual model can achieve state-of-the-art results for this task.

\section{Conclusion and Future Work}
We present the ICDBigBird model, which is a novel contextual model for the ICD coding task. ICDBigBird has the ability to  integrate  a graph embedding model that takes advantage of the relations between ICD codes with a BigBird contextual model that can process larger documents. Experiments on the MIMIC III have shown that the   ICDBigBird model outperforms previous state-of-the-art models. As for future work, we plan to address the limitations of this study including \begin{enumerate*}[label=(\roman*)] \item testing ICDBigBird in other medical datasets to examine its generalizability, strengths and limitations,  \item experimenting on the task of classifying the full ICD code set and \item   examining  the performance of the model  in  datasets of other languages \cite{9099210}. 
\end{enumerate*}

\section{Acknowledgement}
We acknowledge the generous support from  Amazon Research Awards, MITACS Accelerate Grant (\#IT19239), Semantic Health Inc.

\section*{Ethical Consideration}
The ICD coding task is crucial for making clinical, operational and financial decision in healthcare. Traditionally, medical coders review clinical documents and manually assign the appropriate ICD codes by following specific coding guidelines.  Models such as our ICDBigBird could help to reduce time and cost in data extraction and reporting significantly.

However, we need to be aware of the risks of over-relying on any automatic encoding model. No matter how efficient an automatic encoding model is, it is still possible to  misclassify patients' condition with erroneous ICD codes which may affect their treatment. Thus we believe that any automatic encoding model should only be used to assist, not replace the judgement of trained clinical professionals.

\bibliography{anthology,custom}

\begin{thebibliography}{24}
\expandafter\ifx\csname natexlab\endcsname\relax\def\natexlab#1{#1}\fi

\bibitem[{Almagro et~al.(2020)Almagro, Unanue, Fresno, and Montalvo}]{9099210}
Mario Almagro, Raquel~Martínez Unanue, Víctor Fresno, and Soto Montalvo.
  2020.
\newblock \href {https://doi.org/10.1109/ACCESS.2020.2997241} {Icd-10 coding of
  spanish electronic discharge summaries: An extreme classification problem}.
\newblock \emph{IEEE Access}, 8:100073--100083.

\bibitem[{Alsentzer et~al.(2019)Alsentzer, Murphy, Boag, Weng, Jindi, Naumann,
  and McDermott}]{alsentzer-etal-2019-publicly}
Emily Alsentzer, John Murphy, William Boag, Wei-Hung Weng, Di~Jindi, Tristan
  Naumann, and Matthew McDermott. 2019.
\newblock \href {https://doi.org/10.18653/v1/W19-1909} {Publicly available
  clinical {BERT} embeddings}.
\newblock In \emph{Proceedings of the 2nd Clinical Natural Language Processing
  Workshop}, pages 72--78, Minneapolis, Minnesota, USA. Association for
  Computational Linguistics.

\bibitem[{Avati et~al.(2018)Avati, Jung, Harman, Downing, Ng, and
  Shah}]{Avati2018ImprovingPC}
Anand Avati, Kenneth Jung, Stephanie Harman, Lance Downing, A.~Ng, and
  Nigam~Haresh Shah. 2018.
\newblock Improving palliative care with deep learning.
\newblock \emph{BMC Medical Informatics and Decision Making}, 18.

\bibitem[{Cao et~al.(2019)Cao, Wei, Gaidon, Arechiga, and Ma}]{cao2019learning}
Kaidi Cao, Colin Wei, Adrien Gaidon, Nikos Arechiga, and Tengyu Ma. 2019.
\newblock Learning imbalanced datasets with label-distribution-aware margin
  loss.
\newblock In \emph{Advances in Neural Information Processing Systems}.

\bibitem[{Cao et~al.(2020)Cao, Chen, Liu, Zhao, Liu, and
  Chong}]{cao-etal-2020-hypercore}
Pengfei Cao, Yubo Chen, Kang Liu, Jun Zhao, Shengping Liu, and Weifeng Chong.
  2020.
\newblock \href {https://doi.org/10.18653/v1/2020.acl-main.282} {{H}yper{C}ore:
  Hyperbolic and co-graph representation for automatic {ICD} coding}.
\newblock In \emph{Proceedings of the 58th Annual Meeting of the Association
  for Computational Linguistics}, pages 3105--3114, Online. Association for
  Computational Linguistics.

\bibitem[{Chalkidis et~al.(2020)Chalkidis, Fergadiotis, Kotitsas, Malakasiotis,
  Aletras, and Androutsopoulos}]{chalkidis-etal-2020-empirical}
Ilias Chalkidis, Manos Fergadiotis, Sotiris Kotitsas, Prodromos Malakasiotis,
  Nikolaos Aletras, and Ion Androutsopoulos. 2020.
\newblock \href {https://doi.org/10.18653/v1/2020.emnlp-main.607} {An empirical
  study on large-scale multi-label text classification including few and
  zero-shot labels}.
\newblock In \emph{Proceedings of the 2020 Conference on Empirical Methods in
  Natural Language Processing (EMNLP)}, pages 7503--7515, Online. Association
  for Computational Linguistics.

\bibitem[{Devlin et~al.(2019)Devlin, Chang, Lee, and
  Toutanova}]{devlin-etal-2019-bert}
Jacob Devlin, Ming-Wei Chang, Kenton Lee, and Kristina Toutanova. 2019.
\newblock \href {https://doi.org/10.18653/v1/N19-1423} {{BERT}: Pre-training of
  deep bidirectional transformers for language understanding}.
\newblock In \emph{Proceedings of the 2019 Conference of the North {A}merican
  Chapter of the Association for Computational Linguistics: Human Language
  Technologies, Volume 1 (Long and Short Papers)}, pages 4171--4186,
  Minneapolis, Minnesota. Association for Computational Linguistics.

\bibitem[{Farkas and Szarvas(2008)}]{Farkas2008AutomaticCO}
Rich{\'a}rd Farkas and Gy{\"o}rgy Szarvas. 2008.
\newblock Automatic construction of rule-based icd-9-cm coding systems.
\newblock \emph{BMC Bioinformatics}, 9:S10 -- S10.

\bibitem[{Ji et~al.(2020)Ji, Cambria, and Marttinen}]{ji-etal-2020-dilated}
Shaoxiong Ji, Erik Cambria, and Pekka Marttinen. 2020.
\newblock \href {https://doi.org/10.18653/v1/2020.clinicalnlp-1.8} {Dilated
  convolutional attention network for medical code assignment from clinical
  text}.
\newblock In \emph{Proceedings of the 3rd Clinical Natural Language Processing
  Workshop}, pages 73--78, Online. Association for Computational Linguistics.

\bibitem[{Johnson et~al.(2016)Johnson, Pollard, Shen, Li-wei, Feng, Ghassemi,
  Moody, Szolovits, Celi, and Mark}]{mimiciii}
Alistair~EW Johnson, Tom~J Pollard, Lu~Shen, H~Lehman Li-wei, Mengling Feng,
  Mohammad Ghassemi, Benjamin Moody, Peter Szolovits, Leo~Anthony Celi, and
  Roger~G Mark. 2016.
\newblock Mimic-iii, a freely accessible critical care database.
\newblock \emph{Scientific data}, 3:160035.

\bibitem[{Kipf and Welling(2017)}]{kipf2017semisupervised}
Thomas~N. Kipf and Max Welling. 2017.
\newblock \href {https://openreview.net/forum?id=SJU4ayYgl} {{Semi-Supervised
  Classification with Graph Convolutional Networks}}.
\newblock In \emph{Proceedings of the 5th International Conference on Learning
  Representations}, ICLR '17.

\bibitem[{Lee et~al.(2019)Lee, Yoon, Kim, Kim, Kim, So, and
  Kang}]{10.1093/bioinformatics/btz682}
Jinhyuk Lee, Wonjin Yoon, Sungdong Kim, Donghyeon Kim, Sunkyu Kim, Chan~Ho So,
  and Jaewoo Kang. 2019.
\newblock \href {https://doi.org/10.1093/bioinformatics/btz682} {{BioBERT: a
  pre-trained biomedical language representation model for biomedical text
  mining}}.
\newblock \emph{Bioinformatics}, 36(4):1234--1240.

\bibitem[{Li and Yu(2020)}]{lifei}
Fei Li and Hong Yu. 2020.
\newblock \href {https://doi.org/10.1609/aaai.v34i05.6331} {Icd coding from
  clinical text using multi-filter residual convolutional neural network}.
\newblock \emph{Proceedings of the AAAI Conference on Artificial Intelligence},
  34:8180--8187.

\bibitem[{Loshchilov and Hutter(2019)}]{loshchilov2019decoupled}
Ilya Loshchilov and Frank Hutter. 2019.
\newblock Decoupled weight decay regularization.
\newblock In \emph{International Conference on Learning Representation}.

\bibitem[{Lu et~al.(2020)Lu, Du, and Nie}]{lu2020vgcn}
Zhibin Lu, Pan Du, and Jian-Yun Nie. 2020.
\newblock Vgcn-bert: augmenting bert with graph embedding for text
  classification.
\newblock In \emph{European Conference on Information Retrieval}, pages
  369--382. Springer.

\bibitem[{Mullenbach et~al.(2018)Mullenbach, Wiegreffe, Duke, Sun, and
  Eisenstein}]{mullenbach-etal-2018-explainable}
James Mullenbach, Sarah Wiegreffe, Jon Duke, Jimeng Sun, and Jacob Eisenstein.
  2018.
\newblock \href {https://doi.org/10.18653/v1/N18-1100} {Explainable prediction
  of medical codes from clinical text}.
\newblock In \emph{Proceedings of the 2018 Conference of the North {A}merican
  Chapter of the Association for Computational Linguistics: Human Language
  Technologies, Volume 1 (Long Papers)}, pages 1101--1111, New Orleans,
  Louisiana. Association for Computational Linguistics.

\bibitem[{Peters et~al.(2018)Peters, Neumann, Iyyer, Gardner, Clark, Lee, and
  Zettlemoyer}]{peters-etal-2018-deep}
Matthew Peters, Mark Neumann, Mohit Iyyer, Matt Gardner, Christopher Clark,
  Kenton Lee, and Luke Zettlemoyer. 2018.
\newblock \href {https://doi.org/10.18653/v1/N18-1202} {Deep contextualized
  word representations}.
\newblock In \emph{Proceedings of the 2018 Conference of the North {A}merican
  Chapter of the Association for Computational Linguistics: Human Language
  Technologies, Volume 1 (Long Papers)}, pages 2227--2237, New Orleans,
  Louisiana. Association for Computational Linguistics.

\bibitem[{Reimers and Gurevych(2019)}]{sentencebert}
Nils Reimers and Iryna Gurevych. 2019.
\newblock \href {https://doi.org/10.18653/v1/D19-1410} {Sentence-{BERT}:
  Sentence embeddings using {S}iamese {BERT}-networks}.
\newblock In \emph{Proceedings of the 2019 Conference on Empirical Methods in
  Natural Language Processing and the 9th International Joint Conference on
  Natural Language Processing (EMNLP-IJCNLP)}, pages 3982--3992, Hong Kong,
  China. Association for Computational Linguistics.

\bibitem[{Rios and Kavuluru(2018)}]{rios-kavuluru-2018-shot}
Anthony Rios and Ramakanth Kavuluru. 2018.
\newblock \href {https://doi.org/10.18653/v1/D18-1352} {Few-shot and zero-shot
  multi-label learning for structured label spaces}.
\newblock In \emph{Proceedings of the 2018 Conference on Empirical Methods in
  Natural Language Processing}, pages 3132--3142, Brussels, Belgium.
  Association for Computational Linguistics.

\bibitem[{{Shi} et~al.(2017){Shi}, {Xie}, {Hu}, {Zhang}, and
  {Xing}}]{shi2017towards}
Haoran {Shi}, Pengtao {Xie}, Zhiting {Hu}, Ming {Zhang}, and Eric~P. {Xing}.
  2017.
\newblock Towards automated icd coding using deep learning.
\newblock \emph{arXiv preprint arXiv:1711.04075}.

\bibitem[{Wang et~al.(2018)Wang, Li, Wang, Zhang, Shen, Zhang, Henao, and
  Carin}]{wang-etal-2018-joint-embedding}
Guoyin Wang, Chunyuan Li, Wenlin Wang, Yizhe Zhang, Dinghan Shen, Xinyuan
  Zhang, Ricardo Henao, and Lawrence Carin. 2018.
\newblock \href {https://doi.org/10.18653/v1/P18-1216} {Joint embedding of
  words and labels for text classification}.
\newblock In \emph{Proceedings of the 56th Annual Meeting of the Association
  for Computational Linguistics (Volume 1: Long Papers)}, pages 2321--2331,
  Melbourne, Australia. Association for Computational Linguistics.

\bibitem[{Wolf et~al.(2020)Wolf, Debut, Sanh, Chaumond, Delangue, Moi, Cistac,
  Rault, Louf, Funtowicz, Davison, Shleifer, von Platen, Ma, Jernite, Plu, Xu,
  Scao, Gugger, Drame, Lhoest, and Rush}]{Wolf2019HuggingFacesTS}
Thomas Wolf, Lysandre Debut, Victor Sanh, Julien Chaumond, Clement Delangue,
  Anthony Moi, Pierric Cistac, Tim Rault, Rémi Louf, Morgan Funtowicz, Joe
  Davison, Sam Shleifer, Patrick von Platen, Clara Ma, Yacine Jernite, Julien
  Plu, Canwen Xu, Teven~Le Scao, Sylvain Gugger, Mariama Drame, Quentin Lhoest,
  and Alexander~M. Rush. 2020.
\newblock \href {https://www.aclweb.org/anthology/2020.emnlp-demos.6}
  {Transformers: State-of-the-art natural language processing}.
\newblock In \emph{Proceedings of the 2020 Conference on Empirical Methods in
  Natural Language Processing: System Demonstrations}, pages 38--45, Online.
  Association for Computational Linguistics.

\bibitem[{Zaheer et~al.(2020)Zaheer, Guruganesh, Dubey, Ainslie, Alberti,
  Ontanon, Pham, Ravula, Wang, Yang et~al.}]{zaheer2020bigbird}
Manzil Zaheer, Guru Guruganesh, Kumar~Avinava Dubey, Joshua Ainslie, Chris
  Alberti, Santiago Ontanon, Philip Pham, Anirudh Ravula, Qifan Wang, Li~Yang,
  et~al. 2020.
\newblock Big bird: Transformers for longer sequences.
\newblock \emph{Advances in Neural Information Processing Systems}, 33.

\bibitem[{Zhang et~al.(2020)Zhang, Liu, and Razavian}]{zhang-etal-2020-bert}
Zachariah Zhang, Jingshu Liu, and Narges Razavian. 2020.
\newblock \href {https://doi.org/10.18653/v1/2020.clinicalnlp-1.3}
  {{BERT}-{XML}: Large scale automated {ICD} coding using {BERT} pretraining}.
\newblock In \emph{Proceedings of the 3rd Clinical Natural Language Processing
  Workshop}, pages 24--34, Online. Association for Computational Linguistics.

\end{thebibliography}
\bibliographystyle{acl_natbib}

\appendix

\end{document}